\documentclass[conference]{IEEEtran}
\newtheorem{theorem}{Theorem}

\newcommand\blfootnote[1]{%
  \begingroup
  \renewcommand\thefootnote{}\footnote{#1}%
  \addtocounter{footnote}{-1}%
  \endgroup
}

\usepackage{ifpdf}

\ifCLASSOPTIONcompsoc
\usepackage[nocompress]{cite}
\else
\usepackage{cite}
\fi

\ifCLASSINFOpdf
  \usepackage[pdftex]{graphicx}
\else
  \usepackage[dvips]{graphicx}
\fi

\usepackage{amsmath}

\usepackage[ruled,vlined,english,onelanguage]{algorithm2e} 

\usepackage{array}

\ifCLASSOPTIONcompsoc
  \usepackage[caption=false,font=footnotesize,labelfont=sf,textfont=sf]{subfig}
\else
  \usepackage[caption=false,font=footnotesize]{subfig}
\fi

\usepackage{url}

\usepackage{verbatim} 
\usepackage{amssymb} 
\usepackage{caption} 
\usepackage{tikz, tkz-euclide} 
	\tikzset{basic/.style={draw,fill=blue!20}}
	\tikzset{input/.style={basic,circle}}
	\tikzset{weights/.style={basic,rectangle}}
	\tikzset{functions/.style={basic,circle,fill=blue!10}}
	\tikzset{joint/.style={draw,minimum size=0mm}}

\title{A Generalization of Principal Component Analysis \\ { \vspace{-5pt} \large Samuele Battaglino and Erdem Koyuncu$^{\dagger}$ } \vspace{-38pt} }

\begin{document}

\maketitle

\begin{abstract}
Conventional principal component analysis (PCA) finds a principal vector that maximizes the sum of second powers of principal components. We consider a generalized PCA that aims at maximizing the sum of an arbitrary convex function of principal components. We present a gradient ascent algorithm to solve the problem. For the kernel version of generalized PCA, we show that the solutions can be obtained as fixed points of a simple single-layer recurrent neural network. We also evaluate our algorithms on different datasets.\blfootnote{$^{\dagger}$S. Battaglino is with the Department of Computer Science, University of Illinois at Chicago (UIC). E. Koyuncu is with the Department of Electrical and Computer Engineering, UIC. E-mails: \{sbatta2, ekoyuncu\}@uic.edu.}
\end{abstract}

\begin{IEEEkeywords}
Principal component analysis (PCA), Kernel PCA, Recurrent neural networks.
\end{IEEEkeywords}

\maketitle

\IEEEdisplaynontitleabstractindextext

%
\IEEEpeerreviewmaketitle

\section{Introduction}
\label{secintro}
\subsection{Conventional Principal Component Analysis (PCA)}
PCA and variant methods are dimension reduction techniques  that rely on orthogonal transformations \cite{pearson:pca, hotelling:pca, pca:tutorial}. Specifically, let $X = \{x_1,\ldots,x_N\} \subset \mathbb{R}^d$ be a set of $d$-dimensional input vectors with zero mean. In the conventional $L^2$-PCA, one first ``extracts'' the first principal vector via
\begin{align}
\label{l2pcaextract}
w_{L^2}^{(1)} \triangleq \arg\max_{w:\|w\| = 1} \textstyle\sum_{i=1}^N (w^T x_i)^2,
\end{align}
where $(\cdot)^T$ is the matrix transpose, and $\|w\|\triangleq w^T w$ is the $L^2$-norm. The solution of (\ref{l2pcaextract})  is the eigenvector corresponding to the largest eigenvalue of the sample covariance matrix $\sum_{i=1}^N x_i x_i^T$. Subsequent principal vectors $w_{L^2}^{(j)}$, where $2 \leq j \leq \min\{d, N\}$ can then be extracted through the recursions
\begin{align}
\!\!w_{L^2}^{(j)} \triangleq \arg\!\!\max_{\!\!\!\!\!\!w:\|w\| = 1} \textstyle\sum_{i=1}^N \bigl(w^T (I - \sum_{k=1}^{j-1} w_{L^2}^{(k)}(w_{L^2}^{(k)})^T) x_i\bigr)^2\!\!\!.\!\!
\end{align}
In other words, to extract subsequent principal vectors, one projects $x_i$s to the subspace that is orthogonal to the previous principal vectors, and solves (\ref{l2pcaextract}) for the projected $x_i$s. The $j$th principal component of a given input $x_i$ is defined as the magnitude $w_{L^2}^{(j)}x_i$ of $x_i$ along the direction $w_{L^2}^{(j)}$. Dimension reduction is achieved by considering only the first few principal vectors and the corresponding components \cite{intro:dimRed}. PCA can be utilized in a variety of applications including novelty detection \cite{kpca:novelty}, data clustering \cite{intro:clust, koyuncu2017minimum, koyuncu2010systematic, guo2018source}, and outlier detection \cite{intro:outlier, muneeb2019robust}.

\subsection{$L^p$-PCA}
In its original $L^2$-form in (\ref{l2pcaextract}), PCA is very susceptible to outliers or highly-noisy datasets. This is because a noisy sample or an outlier $x_i$ may result in a large inner product $w^T x_i$, which will be further amplified as one considers the second power of  $w^T x_i$ in the objective function. Several variants have thus been proposed to increase the robustness of the conventional $L^2$-PCA, which maximizes the $L^2$-norm of the component vector $[w^Tx_1\,\,w^Tx_2\,\,\cdots\,\,w^Tx_N]$. In particular, one can consider maximizing the $L^p$-norm of the component vector for a general $p$, resulting in the method of $L^p$-PCA \cite{kwak:lppca}. In this case, the first principal vector is extracted via
\begin{align}
\label{lppcaextract}
w_{L^p} \triangleq \arg\max_{w:\|w\| = 1} \textstyle\sum_{i=1}^N |w^T x_i|^p.
\end{align}
The case $p=1$ has been studied in \cite{kwak:l1pca} and is NP-hard \cite{mark:l1np,coy:l1np}. Also, \cite{kwak:lppca} has studied the case of an arbitrary $p$. Once (\ref{lppcaextract}) is solved, subsequent principal vectors can be extracted in a greedy fashion (by projections to orthogonal subspaces as in the conventional $L^2$-PCA). Joint  extraction of multiple principal components (as opposed to greedy extraction) typically provides a better performance \cite{kwak:lppca, intro:nonGreedyL1pca} unless one considers conventional PCA. It has been observed that, in applications to classification, $L^p$-PCA with a general $p \in\{0.5,1.5\}$ can outperform both $L^1$-PCA and $L^2$-PCA for certain datasets \cite{kwak:lppca}. Another robust PCA that provides a rotationally-invariant $L^1$-PCA is \cite{ding2006r}. Further discussions on the robustness advantages of $L^1$-PCA over $L^2$-PCA  can be found in \cite{bacc:l1pca}. We also refer to \cite{lerman2018overview} for a review of $L^1$-PCA and related methods in the lens of robust subspace recovery.

\subsection{Generalized PCA}
\label{genpcasolution}
In this work, we study the generalized PCA problem
\begin{align}
\label{fpcaproblem}
w_f \triangleq \arg\max_{w:\|w\| = 1}\textstyle \sum_{i=1}^N f(w^T x_i),
\end{align}
where $f$ is an arbitrary convex function. The convexity of $f$ is required later so as to guarantee the convergence of algorithms to solve (\ref{fpcaproblem}). Special cases of the formulation (\ref{fpcaproblem}) includes the $L^p$-PCA, which can be recovered by setting $f(x) = |x|^p$. 

The motivation for the generalized PCA in (\ref{fpcaproblem}) is as follows: As discussed above,  $L^1$-PCA and $L^2$-PCA perform the best for data with and without outliers, respectively. However, an outlier typically induces a principal component with a large magnitude. Therefore, for the optimal performance, $f$ should behave as the $L^1$ norm $f(x) \approx x$ for outliers (large $|x|$), and it should behave as the $L^2$ norm $f(x) \approx x^2$ for normalities (small $|x|$). By allowing an arbitrary $f$, we effectively allow a cross between the $L^1$- and $L^2$-norm PCAs (or even $L^p$-PCAs), achieving the best of both worlds.

\subsection{Generalized Kernel PCA (KPCA)}
\label{genkernelpcasolution}
We also study the kernel version of the problem in (\ref{fpcaproblem}). In other words, we consider the problem
\begin{align}
\label{gkpcaproblem}
w_{f, \Phi} \triangleq \arg\max_{w:\|w\| = 1} \textstyle\sum_{i=1}^N f(w^T \Phi(x_i)),
\end{align}
where $\Phi(\cdot)$ is an arbitrary feature map. We present a solution to (\ref{gkpcaproblem}) in terms of a simple recurrent neural network. Our solution does not need the computation of the feature maps $\Phi(x_i),\,i=1,\ldots,N$. Special cases of our algorithm include algorithms to solve the conventional KPCA \cite{scholkopf:kpca}, and the $L^1$-KPCA \cite{kim:l1kpca}. Similar connections between neural networks and PCA have also been previously established in the literature \cite{intro:nlpca,intro:classifier,intro:kpcaLearn}. 

\subsection{Organization of the Paper}
The rest of the paper is organized as follows: In Section \ref{secalgorithms}, we present our algorithms to solve the generalized PCA and KPCA problems in (\ref{fpcaproblem}) and (\ref{gkpcaproblem}). In Section \ref{secnumerical}, we present numerical results over different datasets. Finally, in Section \ref{secconclusions}, we draw our main conclusions.

\section{Algorithms for the Generalized PCA Problems}
\label{secalgorithms}
\subsection{An Algorithm for Generalized Non-Kernel PCA}
First, we focus on the solution of the generalized PCA problem as stated in (\ref{fpcaproblem}). We will utilize the following result.
\begin{theorem}[\!\!{\cite[Theorem 1]{kwak:lppca}}]
\label{theoremi}
Let $F(w)$ be a convex function. Let $\|w'\| = 1$ and $w'' = \frac{\nabla F}{\|\nabla F\|}|_{w = w'}$. Then, $F(w') \geq F(w)$. In particular, the gradient ascent $w \leftarrow  \frac{\nabla F}{\|\nabla F\|}$ provides a locally-maximum solution to the problem $\max_{w:\|w\| = 1} F(w)$.
\end{theorem}

In order to apply Theorem \ref{theoremi} to solve (\ref{fpcaproblem}), we first note that $w \mapsto f(w^T x_i)$ is convex for any $x_i$, as the composition of a convex function and an affine function is always convex. It follows that $w \mapsto \sum_{i=1}^N f(w^T x_i)$ is convex and Theorem \ref{theoremi} is applicable. We may thus use the gradient ascent
\begin{align}
\label{gradascentrule}
w \leftarrow \frac{\sum_{i=1}^N f'(w^T x_i)x_i }{\|\sum_{i=1}^N f'(w^T x_i)x_i\|}
\end{align}
to solve (\ref{fpcaproblem}) and extract the first principal vector. Subsequent principal vectors are extracted in a greedy fashion as in the conventional PCA. Note that the ascent is not guaranteed to find the globally-optimal solution to (\ref{fpcaproblem}) due to the existence of several local maxima. An important question in this context is how to initialize the ascent. Following the advice in \cite{kwak:lppca}, in our numerical experiments, we have chosen the initial $w$ to be in the same direction as the data vector with the largest $L^2$ norm. In other words, we initialize $w \leftarrow \arg\max_i x_i/\|x_i\|$.  
\subsection{An Algorithm for Generalized KPCA}

\begin{figure}[h]
	\centering
	\scalebox{0.4}{\includegraphics{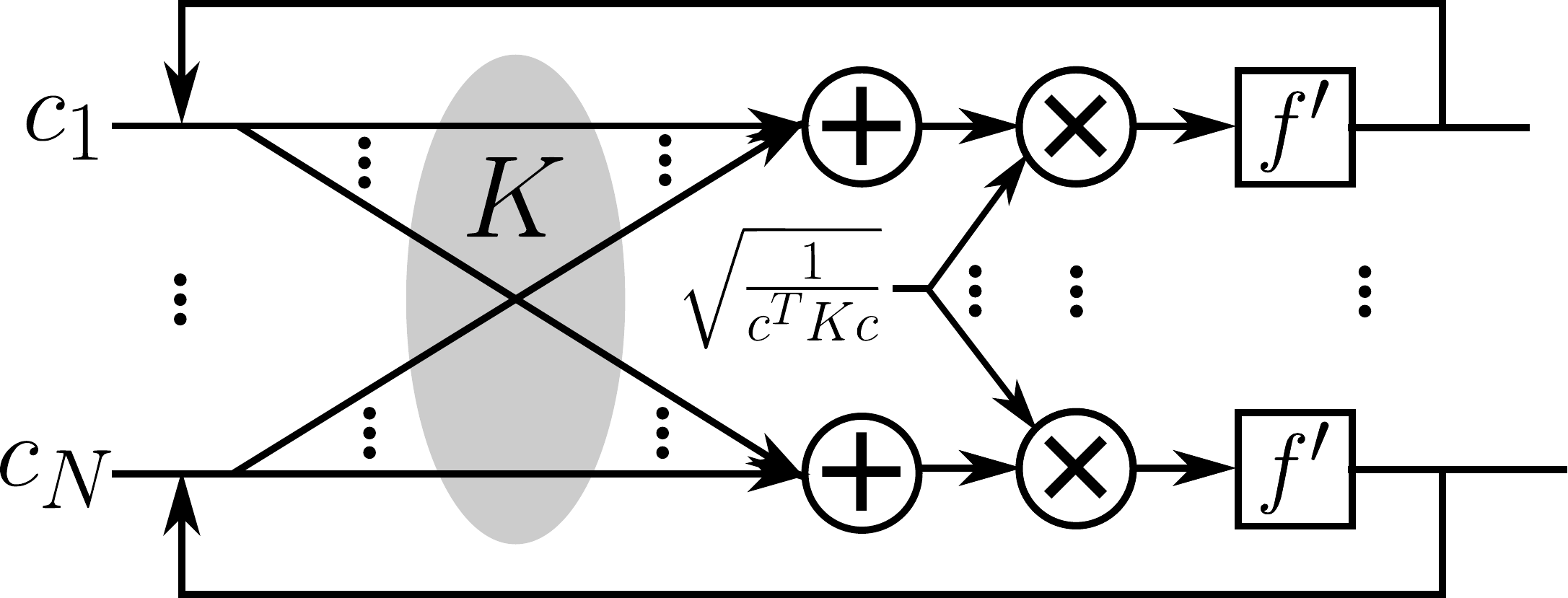}}
	\caption{The recurrent neural network for generalized KPCA.}
	\label{gkpcarnnfig}
\end{figure}

We now consider the solution to the generalized KPCA problem in (\ref{gkpcaproblem}). A simple extension of (\ref{gradascentrule}) results in the gradient ascent algorithm
\begin{align}
\label{gkpca0}
w \leftarrow \frac{\sum_{i=1}^N f'(w^T \Phi(x_i))\Phi(x_i) }{\|\sum_{i=1}^N f'(w^T \Phi(x_i))\Phi(x_i)\|}.
\end{align}
The drawback of this algorithm is that it requires us to compute the feature vectors, which may be infeasible. We thus apply a variant of the ``Kernel trick'' as follows: Let $c_i \triangleq f'(w^T \Phi(x_i))$, and $c = [c_1 \cdots c_N]^T$. Also, let $K$ define the Kernel matrix with entry $(\Phi(x_i))^T \Phi(x_j)$ in the $i$th row, $j$th column. The update rule in (\ref{gkpca0}) can then be written as
\begin{align}
\label{gkpca1}
w \leftarrow \frac{\sum_{j=1}^N c_j \Phi(x_j) }{\sqrt{c^T K c}}.
\end{align}
We now first take the transpose of both sides, then multiply both sides by $(\Phi(x_i))^T$, and then apply the function $f'(\cdot)$ to both sides. This yields the alternative update equations
\begin{align}
\label{gkpca2pre}
c_i \leftarrow f'\left(\frac{\sum_{j=1}^N K_{ij}c_j}{\sqrt{c^T K c}} \right),\,i=1,\ldots,N.
\end{align}
Equivalently, we may write (\ref{gkpca2pre}) in its simpler matrix form
\begin{align}
\label{gkpca2}
c \leftarrow f'\left(\frac{Kc}{\sqrt{c^T K c}} \right),
\end{align}
with the understanding that the function $f'$ is applied component-wise. Note that (\ref{gkpca2}) describes a simple recurrent network to find $c$ and thus extract the principal components, as illustrated in Fig. \ref{gkpcarnnfig}.  Unfortunately, the convergence properties of (\ref{gkpca1}) cannot be directly extended to (\ref{gkpca2}) as the transformations we have used to derive (\ref{gkpca2}) from (\ref{gkpca1}) are not one to one. In fact, a full analysis of the iterative algorithm (\ref{gkpca2}) for a general $f$ appears to be a difficult problem. Nevertheless, in practical datasets, we have always observed (\ref{gkpca2}) to converge, so that such an analysis remains more of a theoretical curiosity.

It is also worth mentioning that convergence of (\ref{gkpca2}) can be analytically established for certain special cases. In particular for conventional  KPCA, where we set $f(x) = x^2$, the iterations (\ref{gkpca2}) are equivalent to power method iterations \cite{bjorck2015numerical}. In this case, it is well-known that the vector $c$ converges to the eigenvector corresponding to the largest eigenvalue of $K$.

Another special case is when $f(x) = |x|$ corresponds to $L^1$-KPCA. In this case, $f'(x) = \mathrm{sign}(x)$, and the iterations in (\ref{gkpca2}) simplify to $c \leftarrow \mathrm{sign}(Kc)$. This describes, in fact, the parallel update rule in a Hopfield network \cite{hop:hopnet}. According to \cite{bruck1990convergence}, the iterations then converge to a cycle of length $2$ in general. Again, in practice, we have always observed convergence without any cyclic behavior. To guarantee convergence, one can implement $c \leftarrow \mathrm{sign}(Kc)$ with serial, instead of parallel operation. In this case, the components of $c$ are updated one at a time instead of all simultaneously  \cite{bruck1990convergence}. With serial operation, the iteration is guaranteed to converge to a unique $c$. The serial implementation of the iterations $c \leftarrow \mathrm{sign}(Kc)$ yields the exact same algorithm that is used in \cite{kim:l1kpca} to solve the $L^1$-KPCA problem. Our work thus provides an alternative derivation and shows that Hopfield networks can also be understood as principal component analyzers.

Let us now proceed with the assumption that (\ref{gkpca2}) converges. Knowing $c$, the principal component $w^T x$ for any given input $x$ can then calculated via (\ref{gkpca1}) combined with the Kernel trick. Specifically, we have $w^T x = \sum_{j=1}^N c_j K(x_j, x) / \sqrt{c^T K c}$. Other implementation aspects of the conventional KPCA \cite{scholkopf:kpca} extend in a straightforward manner. For example, after the construction of the kernel matrix $K$, the features are normalized to have zero mean via the transformation $K \leftarrow K - YK - KY + YKY$, where $Y$ represents the all-$1$ matrix. Also, to greedily calculate the second principal vector, we need to find the kernel matrix corresponding to the inputs $(I - ww^T)x_i$, where $w$ is given by (\ref{gkpca1}). 
Straightforward calculations show that the corresponding update on the kernel matrix is given by $K \leftarrow K - K c c^T K / c^T K c$. 

Another issue is how to choose the initial $c$. Following the methodology in Section \ref{genpcasolution}, we choose $c$ in such a way that it will induce a principal vector with the same direction as $\arg\max_{x_i} \|\Phi(x_i)\|$. According to (\ref{gkpca1}), this can be accomplished by choosing $c_j = 1$ and $c_i = 0,\,i \neq j$, where the index $j$ satisfies $j = \arg\max_{i} \|\Phi(x_i)\| = \arg\max_i K_{ii}$.

The entire generalized KPCA algorithm is summarized in Algorithm \ref{algogkpca}. As mentioned, special cases are the $L^1$-KPCA \cite{kim:l1kpca} and the $L^2$-KPCA \cite{scholkopf:kpca} algorithms. In particular, Algorithm \ref{algogkpca} also yields an $L^p$-KPCA algorithm for a general $p$ not necessarily equal to $1$ or $2$.

\begin{algorithm}[h!]
	\caption{The Generalized KPCA Algorithm}\label{alg:gkpca}
	\SetAlgoLined
	
	Let $K$ have entry $(\Phi(x_i))^T\Phi(x_j)$ in Row $i$, Column $j$. \\ 
	$K \leftarrow K- Y K-K Y + Y K Y$.\\
	\For{$i\gets1$ \KwTo \#(Principal Vectors to be Extracted)}{
		Set $c_j = 1, c_i = 0,\,i \neq j$, where $j = \arg\max_i K_{ii}$. \\
		Iterate $c \leftarrow f'(\frac{Kc}{\sqrt{c^T K c}} )$ until convergence. \\
		Store $c,K$ to analyze the $i$th principal component. \\
		$K\leftarrow K- K c c^T K / c^T K c$
	}
	\label{algogkpca}
\end{algorithm}




\section{Numerical Results}
\label{secnumerical}
In this section, we present numerical results that demonstrate the performance of our generalized PCA algorithms over existing methods. A key design choice in testing our algorithms is the function $f$. Maximizing the performance with respect to $f$ is an infinite-dimensional optimization problem as $f$ can be chosen to be any convex function. This is why, in the current work, we optimize $f$ over only a certain class of functions, leaving a full optimization for future work. In all experiments, we extract $30$ principal vectors per class. 

First, we present the results for generalized non-kernel PCA. As discussed in Section \ref{secintro}, a major motivation for our generalization of PCA is the individual optimality of $L^1$-PCA and $L^2$-PCA methods in different regimes of the principal component magnitude. Hence, one simple choice can be
\begin{align}
f(x) = g_a(x) \triangleq \left\{\begin{array}{rl} x^2, & x \leq a, \\ |x|, & x > a, \end{array} \right.
\end{align}
where $a > 0 $ is a parameter. The function $g_a$ acts as the $L^2$ and $L^1$ norms for small and large arguments, respectively. We have also considered the two continuous functions
\begin{align}
f'(x) = \zeta_1(x) & \triangleq (1-\mathrm{sech}(|x|))\mathrm{sign}(x), \\
f'(x) = \zeta_2(x) & \triangleq \tanh^2(|x|)\mathrm{sign}(x)
\end{align}
with a similar behavior as $g_a$. 




In Table \ref{tablei}, we show the classification performance of different generalized PCA (GPCA) methods over the USPS dataset  \cite{roweis:usps} with different Gaussian noise variances $\sigma$. Shown numbers are the classification accuracies obtained as follows: As mentioned before, $30$ principal vectors are extracted per class. Given a test sample, the class whose principal vectors provide the lowest reconstruction error is chosen and compared with the desired output. In Table \ref{tableii}, we show the results with the salt and pepper (S\&P) noise. The probability that a pixel flips from black to white or white to black is given by $\frac{\delta}{2}$. For both types of noise, we can observe that generalized PCA outperforms with non-$L^p$-norm functions outperform $L^p$-PCA methods for every $p\in\{0.5,1,1.5,2\}$. The function $\zeta_1$ is particularly effective at high noise levels. 

In Table \ref{tableiii}, we show the percent misclassification rates for the MNIST dataset \cite{lecun:mnist} with different speckle noise variances $\eta$. In this case, $L^1$-PCA also proves to be an effective method at low noise. The crossover $g_1$  between $L^1$-PCA and $L^2$-PCA provides excellent performance at all noise levels. 

We also present results on the Yale Faces dataset \cite{yale:faces, yale:mit}, which consists of $165$ images of several individuals. Following the testing methodology in \cite{kwak:lppca}, we add a number of noise images consisting of random black and white pixels only. We compute the principal vectors on the extended dataset including these noise images. We report the $L^2$-reconstruction errors (scaled down by a factor of $1000$) for the original dataset of $165$ images in Table \ref{tableiv}. We can observe that generalized PCA with the choice $f' = \zeta_2$ provides the best performance, especially in the presence of noise. In the noiseless case, conventional $L^2$-PCA provides the best reconstruction error.



We now provide numerical results for our generalized KPCA method. In our experiments, we used the Gaussian kernel $(\Phi(x_i))^T\Phi(x_j)=\exp(\frac{1}{\rho^2}||x_1-x_2||^2)$, where $\rho > 0$ is a parameter. In our experiments, we performed an exhaustive search to optimize $\rho$ for $L^2$-KPCA only. The resulting $\rho$ is then kept fixed for the different functions $f$ we evaluate.

In regular PCA, while choosing the functions $f$, we frequently referred to our intuition that large principal components most likely correspond to outliers/noisy samples. On the other hand, for kernel PCA, since the dataset is already altered by the feature map of the kernel, this earlier intuition may not be valid. In fact, as we shall demonstrate, the family of Gaussian-like functions 
\begin{align}
\label{fprimekernel}
f'(x) = h_q(x) \triangleq e^{-|x|^q} \mathrm{sign}(x),
\end{align}
which are parameterized by some $q > 0$, provides the best performance. We can imagine that the family is matched to the kernel due to its similarity to the Gaussian function. Also, although $f$ is no longer convex in (\ref{fprimekernel}), we observed convergence for all instances of experiments.


\begin{table}[!t]
    		\centering
    		\captionof{table}{PCA - USPS WITH GAUSSIAN NOISE} 
    		\begin{tabular}{ c | l c l c c c}
    			\label{table:GuspsGaus}
    			$\sigma$ & $p=0.5$ & $p=1$  & $p=1.5$ &  $p=2$ & $\zeta_1$ & $g_1$ \\
    			\hline
    			0  & 95.08 & 95.52 & 95.51 & 95.46  & 95.35 & \textbf{95.65} \\
    			10 & 95.02 & 95.42 & 95.45 & 95.45  & 95.34 & \textbf{95.55} \\
    			20 & 94.89 & 95.31 & 95.34 & 95.31 &  95.31 & \textbf{95.41} \\
    			30 & 94.71 & 95.14 & 95.17 & 95.17 &  95.19 & \textbf{95.25} \\
    			40 & 94.47 & 94.86 & 94.93 & 94.94 &  95.00 & \textbf{95.02} \\
    			50 & 94.11 & 94.52 & 94.60 &  94.60 & 94.58 & \textbf{94.72} \\
    			60 & 93.64 & 94.01 &  94.13 &  94.13 & \textbf{94.29} & 94.27 \\
    			70 & 92.99 & 93.42 &  93.46 &  93.39  & \textbf{93.68} & 93.62 \\
    			80 & 92.00 & 92.49 &  92.53 &  92.31  & \textbf{92.82} & 92.69 \\
    			90 & 90.67 & 91.20 &  91.15 & 90.82  & \textbf{91.56} & 91.40 \\
    			100 & 88.76 & 89.47 &  89.26 &  88.75  & \textbf{89.83} & 89.58 \\
    			\hline
    			Average & 93.30 & 93.76 &  93.78 &  93.67 & 93.91 & \textbf{93.92} \\
    		\end{tabular}
    		\label{tablei}
    	\end{table}
    	
    	\begin{table}[!t]
    		\centering	
    		\captionof{table}{PCA - USPS WITH S\&P NOISE}
    		\begin{tabular}{ c | l c l c c c}
    			\label{table:GuspsSP}
    			$\delta$ & $p=0.5$ & $p=1$  & $p=1.5$ &  $p=2$ & $\zeta_1$ & $g_1$ \\
    			\hline
    			0    & 95.08 & 95.52 & 95.51 & 95.46 & 95.35 & \textbf{95.65} \\
    			0.05 & 94.36 & 94.79 & 94.92 & 94.90 & \textbf{94.95} & 94.91 \\
    			0.1  & 93.36 & 93.76 & 93.84 & 93.87 & \textbf{93.95} & 93.91 \\
    			0.15 & 91.78 & 92.17 & 92.23 & 92.24 & 92.30 & \textbf{92.38} \\
    			0.2  & 89.45 & 89.78 & 89.75 & 89.77 & 89.83 & \textbf{90.08} \\
    			0.25 & 86.08 & 86.37 & 86.25 & 86.19 & 86.37 & \textbf{86.82} \\
    			0.3  & 81.74 & 82.17 & 81.98 & 81.24 & \textbf{82.69} & 82.25 \\
    			0.35 & 76.08 & 76.43 & 75.89 & 74.95 & \textbf{76.95} & 76.45 \\
    			0.4  & 69.28 & 69.60 & 68.76 & 67.54 & \textbf{69.95} & 69.51 \\
    			0.45 & 61.71 & 61.98 & 61.00 & 59.52 & \textbf{62.33} & 61.73 \\
    			0.5  & 53.96 & 53.94 & 52.91 & 51.22 & \textbf{54.22} & 53.68 \\
    			\hline
    			Average & 81.17 & 81.50 &  81.18 &  80.63 & \textbf{81.72} & 81.58 \\
    		\end{tabular}
    		    		\label{tableii}
    	\end{table}
    	
    	\begin{table}[!t]
    		\centering	
    		\captionof{table}{PCA - MNIST WITH SPECKLE NOISE}
    		\begin{tabular}{ c | l c l c c c}
    			\label{table:Gmnist}
    			$\eta$ & $p=0.5$ & $p=1$  & $p=1.5$ &  $p=2$ & $\zeta_1$ & $g_1$ \\
    			\hline
    			0  & 4.22 & \textbf{4.17} & 4.28 & 4.28  & \textbf{4.17} & \textbf{4.17} \\
    			1  & 7.12 & \textbf{6.86} & 7.02 & 8.65 & 6.90 & 6.87 \\
    			2  & 8.32 & 8.07 & 8.19 & 8.33 & 8.07 & \textbf{8.04} \\
    			4  & 9.45 & 9.22 & 9.39 & 9.52 & 9.17 & \textbf{9.16} \\
    			5  & 9.82 & 9.56 & 9.72 & 9.86 & \textbf{9.52} & 9.53 \\
    			8  & 10.41 & 10.25 & 10.43 & 10.59 & 10.23 & \textbf{10.19} \\
    			10 & 10.79 & 10.56 & 10.73 & 10.91 & 10.51 & \textbf{10.48} \\
    			\hline
    		Average & 8.59 & 8.38 & 8.54 & 8.88 & 8.37 & \textbf{8.35} \\
    		\end{tabular}
    		    		\label{tableiii}
    	\end{table}
    	
    	\begin{table}[!t]
    		\centering	
    		\captionof{table}{PCA - YALE FACES}
    		\begin{tabular}{ c | l c l c c c}
    			\label{table:Gyale}
    			$\#$(noisy images) & $p$=0.5 & $p$=1  & $p$=1.5 &  $p$=2 & $\zeta_1$ & $\zeta_2$ \\
    			\hline
    			0 (0\%)  & 2.02 & 1.82 & 1.76 & \textbf{1.75} & 1.82 & 1.83\\    
    			15 (9\%)  & 2.10 & 2.07 & 2.13 & 2.24  & 2.05 & \textbf{2.04}   \\ 
    			30 (18\%)  & 2.20 & \textbf{2.18} & 2.31 & 2.64 & \textbf{2.18} & \textbf{2.18} \\ 
    			45 (27\%)  & 2.28 & 2.29 & 2.43 & 2.79 & 2.28 & \textbf{2.17}  \\ 
    			\hline
    			Average & 2.15 & 2.09  & 2.16 & 2.35  & \textbf{2.08} & \textbf{2.08} \\ 
    		\end{tabular}	
    		    		\label{tableiv}
    	\end{table}

In Table \ref{table:KuspsGaus}, we show the accuracies for the USPS dataset with Gaussian noise and different generalized KPCA functions $f$. We can observe that the accuracies are much higher as compared with the non-kernel results for the same dataset in Table \ref{table:GuspsGaus}. Both $L_2$-KPCA and the generalized KPCA with $f'=g_1$ perform well at low noise levels. On the other hand, at high noise levels, both choices $f' = h_2$ and $f' = h_3$ outperform the existing KPCA methods, with the choice $f' = h_3$ performing the best. Note that the results for $L^{0.5}$-KPCA are also new, as there is no existing $L^p$-KPCA method in the literature unless $p\in\{1,2\}$. As shown in Table \ref{table:KuspsSP}, we can observe that $L^{0.5}$-KPCA performs the second best for high salt and pepper noise. Nevertheless, the choice $f' =  h_3$ still outperforms all other choices for $f'$. We can also observe that the choice $f' = \zeta_1$ performs the worst among all other possible choices for almost all noise levels. This is in contrast to the high performance of $f' = \zeta_1$ in the non-kernel case in Table \ref{tableii}. Different $f$ may thus be optimal in the kernel and non-kernel counterparts of the same PCA problem.

Finally, in Tables \ref{table:Kmnist} and \ref{table:Kyale}, we show the KPCA performance for the MNIST and Yale Face datasets. We can observe that in both cases, $L^{1.5}$-KPCA and KPCA with the choice $f' = h_3$ provides the best performances. 


\begin{table}[!t]
	\centering
	\captionof{table}{KPCA - USPS WITH GAUSSIAN NOISE} 
	\begin{tabular}{ c | c c c c c c}
		\label{table:KuspsGaus}
		$\sigma$ & $p=1$ & $p=2$ & $p=0.5$ & $h_3$ & $h_2$ & $g_1$ \\
		\hline
		0  & 96.26 & \textbf{96.53} & 95.73 & 96.43 & 96.41 & \textbf{96.53} \\
		10 & 96.28 & \textbf{96.50} & 95.69 & 96.45 & 96.35 & 96.49 \\
		20 & 96.27 & \textbf{96.46} & 95.71 & 96.40 & 96.34 & 96.45 \\
		30 & 96.24 & \textbf{96.40} & 95.66 & 96.34 & 96.28 & 96.39 \\
		40 & 96.18 & \textbf{96.30} & 95.60 & 96.24 & 96.20 & 96.29 \\
		50 & 96.08 & \textbf{96.09} & 95.43 & 96.08 & 95.95 & \textbf{96.10} \\
		60 & 95.81 & 95.79 & 95.20  & 95.81 & 95.70 & \textbf{95.84} \\
		70 & 95.35 & 95.35 & 94.79 & \textbf{95.39} & 95.26 & 95.35 \\
		80 & 95.62 & 94.61 & 94.16 & \textbf{94.69} & 94.63 & 94.60 \\
		90 & 93.54 & 93.47 & 93.21 & \textbf{93.68} & 93.48 & 93.46 \\
		100 & 91.98 & 91.80 & 91.81 & \textbf{92.18} & 91.94 & 91.83 \\
		\hline
		Average & 95.33 & 95.39 & 94.82 & \textbf{95.43} & 95.32 & 95.40\\
	\end{tabular}
\end{table} 

\begin{table}[!t]
	\centering
	\captionof{table}{KPCA - USPS WITH S\&P NOISE}
	\begin{tabular}{ c | c c c c c c}
		\label{table:KuspsSP}
		$\delta$ & $p=1$ & $p=2$ & $p=0.5$ & $h_3$ & $h_2$ & $\zeta_1$ \\
		\hline
		0    & 96.26 & \textbf{96.53} & 95.73 & 96.43 & 96.41 & 96.29 \\
		0.05 & 96.07 & \textbf{96.25} & 95.56 & 96.19 & 96.02 & 96.00 \\
		0.1  & 95.65 & \textbf{95.79} & 95.15 & 95.74 & 95.69 & 95.53 \\
		0.15 & 94.83 & \textbf{94.94} & 94.36 & 94.93 & 94.76 & 94.65 \\
		0.2  & 93.36 & 93.40 & 93.00 & \textbf{93.51} & 93.32 & 93.18 \\
		0.25 & 90.94 & 90.84 & 90.79 & \textbf{91.14} & 90.93 & 90.65 \\
		0.3  & 87.19 & 86.83 & 87.08 & \textbf{87.45} & 87.11 & 86.81 \\
		0.35 & 81.89 & 81.21 & 81.92 & \textbf{82.14} & 81.74 & 81.28 \\
		0.4  & 75.22 & 74.20 & 75.25 & \textbf{75.46} & 75.03 & 74.31 \\
		0.45 & 67.67 & 66.44 & 67.69 & \textbf{67.92} & 67.24 & 66.35 \\
		0.5  & 59.74 & 58.41 & 59.78 & \textbf{60.05} & 59.48 & 58.16 \\
		\hline
		Average & 85.35 & 84.99 &  85.12 & \textbf{85.54} & 85.25 & 84.84\\
	\end{tabular}
\end{table} 

\begin{table}[!t]
	\centering
	\captionof{table}{KPCA - MNIST WITH SPECKLE NOISE}
	\begin{tabular}{ c | c c c c c c}
		\label{table:Kmnist}
		$\eta$ & $p=1$ & $p=2$ & $p=1.5$ & $h_3$ & $h_2$ & $g_1$ \\
		\hline
		0  & 3.62 & 3.61 & 3.65 & 3.74 & 3.76 & \textbf{3.59} \\
		1  & 5.63 & 5.55 & 5.57 & \textbf{5.54} & 5.56 & 5.71 \\
		2  & \textbf{6.41} & 6.57 & \textbf{6.41} & \textbf{6.41} & 6.53 &  6.61 \\
		4  & 7.21 & 7.32 & 7.37 & \textbf{7.19} & 7.22 &  7.36 \\
		5  & 7.46 & 7.71 & 7.58 & \textbf{7.43} & 7.48 & 7.57 \\
		8  & 8.01 & 8.13 & \textbf{7.94} & 7.95 & 7.99 & 8.22 \\
		10 & 8.14 & 8.37 & \textbf{8.11} & \textbf{8.11} & 8.12 &  8.40 \\
		\hline
		Average & 6.64 & 6.75 & 6.66 & \textbf{6.62} & 6.66 &  6.78 \\
	\end{tabular}
\end{table} 

\begin{table}[!t]
	\centering
	\captionof{table}{KPCA - YALE FACES}
	\begin{tabular}{ c | c c c c c c}
		\label{table:Kyale}
		\#(noisy images) & $p=1$ & $p=2$ & $p=1.5$ & $h_3$ & $h_2$ & $\zeta_2$\\
		\hline
		0 (0\%)  & 0.91 & \textbf{0.83} & 0.85 & 0.90 & 0.91 & 0.93\\    
		15 (9\%)  & 1.02 & 1.18 & \textbf{1.01} & 1.02 & 1.03 & 1.45 \\ 
		30 (18\%)  & 1.09 & 1.32 & \textbf{1.08} & \textbf{1.08} & \textbf{1.08} & 2.01\\ 
		45 (27\%)  & 1.14 & 1.35 & \textbf{1.12} & 1.13 & 1.14 & 2.20\\ 
		\hline
		Average & 1.04 & 1.17  & \textbf{1.02} & 1.03 & 1.04 & 1.65\\ 
	\end{tabular}	
\end{table}

\section{Conclusions}
\label{secconclusions}
We have presented a generalized PCA that focuses on maximizing the sum of an arbitrary convex function of principal components. We have presented a simple gradient ascent algorithm for the non-kernel version of the problem, and devised a simple recurrent neural network to solve the general case. We have observed that our generalized PCA outperforms the existing $L^p$-norm based PCA methods in several scenarios.

\renewcommand{\IEEEbibitemsep}{0pt plus 0.5pt}
\makeatletter
\IEEEtriggercmd{\reset@font\normalfont\fontsize{10pt}{12pt}\selectfont}
\makeatother
\IEEEtriggeratref{1}

\bibliographystyle{IEEEtran}
\bibliography{./bibtex/IEEEabrv,./bibtex/bibl}

\end{document}